\documentclass[letterpaper, 10 pt, conference]{ieeeconf}

\IEEEoverridecommandlockouts                            

\usepackage[utf8]{inputenc} 
\usepackage[T1]{fontenc}    
\usepackage{hyperref}       
\usepackage{url}            
\usepackage{booktabs}       
\usepackage{nicefrac}       
\usepackage{microtype}      

\usepackage{siunitx} 

\usepackage{cite}
\usepackage{amsmath,amssymb}
\usepackage{algorithmic}
\usepackage{graphicx}
\usepackage{subfig}
\usepackage{textcomp}
\usepackage{gensymb}
\usepackage{epsfig} 
\usepackage{xcolor}

\usepackage{booktabs}
\usepackage{array}
\usepackage{multirow}

\title{\LARGE \bf Jammkle: Fibre jamming 3D printed multi-material tendons and their application in a robotic ankle}

\author{James Brett$^{1}$*, Katrina Lo Surdo$^{1}$*, Lauren Hanson$^{1}$*, Joshua Pinskier$^{1}$*, and David Howard$^{1}$

\thanks{$^{1}$ Robotics and Autonomous Systems Group, Data61, CSIRO, Australia; contact {david.howard@csiro.au}
* These authors contributed equally to the work.
}

}

\begin{document}
\maketitle
\thispagestyle{empty}
\pagestyle{empty}

\begin{abstract}

Fibre jamming is a relatively new and understudied soft robotic mechanism that has previously found success when used in stiffness-tuneable arms and fingers.  However, to date researchers have not fully taken advantage of the freedom offered by contemporary fabrication techniques including multi-material 3D printing in the creation of fibre jamming structures.  In this research, we present a novel, modular, multi-material, 3D printed, fibre jamming tendon unit for use in a stiffness-tuneable compliant robotic ankle, or \emph{Jammkle}.  We describe the design and fabrication of the Jammkle and highlight its advantages compared to examples from modern literature.  We develop a multiphysics model of the tendon unit, showing good agreement with experimental data. Finally, we demonstrate a practical application by integrating multiple tendon units into a robotic ankle and perform extensive testing and characterisation.  We show that the Jammkle outperforms comparative leg structures in terms of compliance, damping, and slip prevention. 

\end{abstract}


\section{Introduction}

Legged locomotion in rough terrain remains a challenging robotics research problem with potential applications across a gamut of high-impact domains ranging from disaster response \cite{8206364} to biosecurity \cite{jurdak2015autonomous}.  There are two broad approaches to solving this problem: via software and hardware.  In software, terrain-sensing and energetics-based approaches\cite{kottege2015energetics}, together with simulated curriculum learning have shown promise in generating locomotion policies for successful field deployment on physical platforms\cite{lee2020learning,kumar2021rma}.  These approaches often require extensive pre-training, extra sensors, or an accurate simulation model, which may not always be readily available.  Additionally, they can create complex learned controllers.  

Hardware ankle structures across the vast majority of legged robots are rigid and lack compliance, which can lead to jolts and slips (which must be overcome in software) and may damage to the robot.  An attractive alternative leverages notions of {\em material computation} (MC) and {\em embodied cognition} (EC) \cite{pfeifer2009morphological,howard2019evolving} to embed some of the task-solving ability into the morphological design of the robot, allowing more complex tasks to be solved in hardware with simpler controllers. \cite{nygaard2021real}.

\begin{figure}[t!]
\centering
\includegraphics[width=0.8\columnwidth]{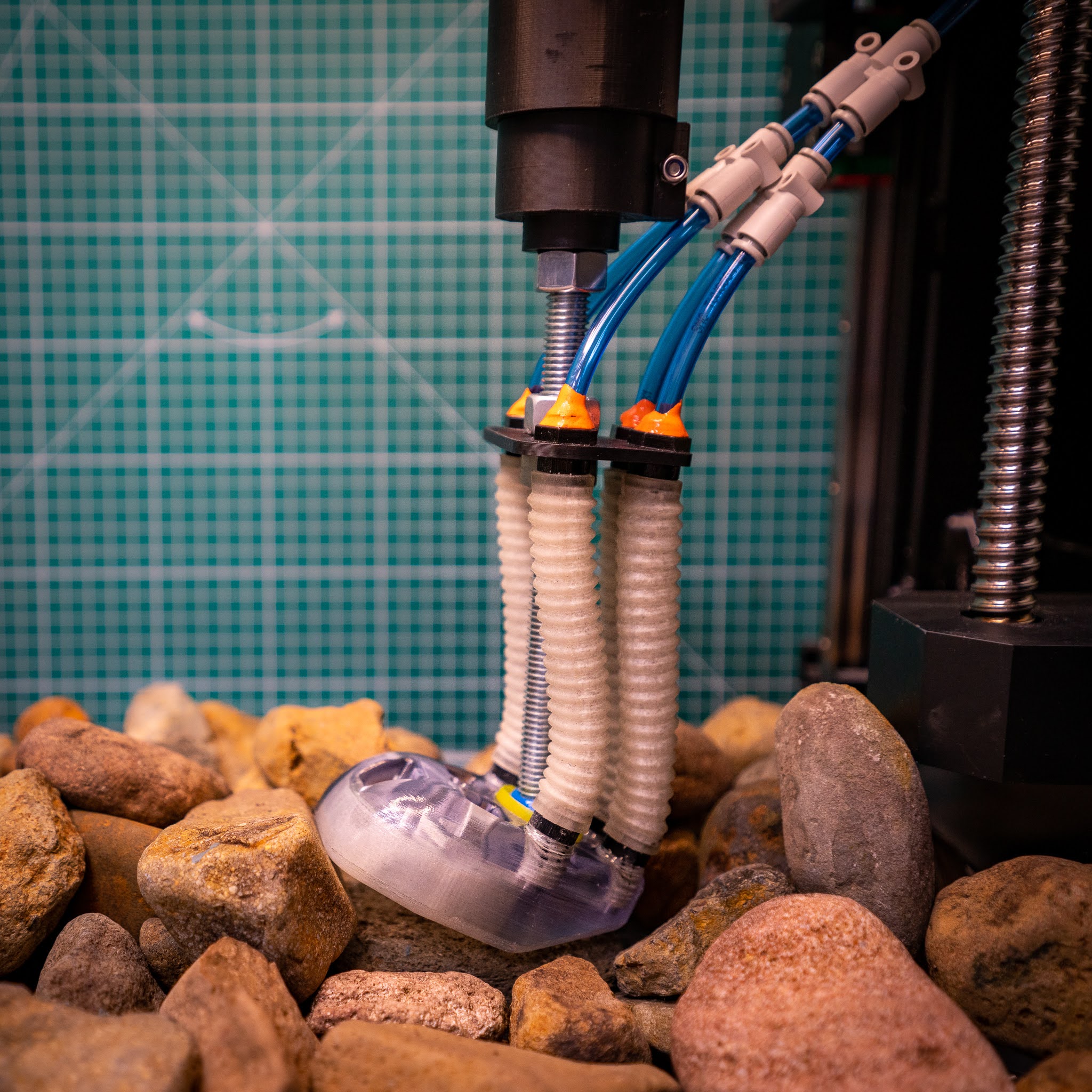}
\caption{Four tendon units integrated into a robotic ankle, showing compliance over rough terrain.}
\label{fig:fig1}
\end{figure}

Soft robotics\cite{shintake2018soft} delivers stiffness-tuneable hardware solutions \cite{7565718} through MC and EC-inspired mechanisms.  Fibre jamming \cite{fitzgerald2020review} is a particularly promising (and understudied) soft robotic mechanism based on the compression of fibrous strands, typically under vacuum pressure.  Fibre jamming has started to appear across a range of robotics applications, mainly as robotic fingers and other end effectors \cite{brancadoro2019toward}.

In this work we develop a novel, modular, 3D printed, multi-material, fibre jamming actuator for use as a compliant tendon.  Multi-material printing provides varied material properties in each individual fibre of the jamming structure, promoting excellent stiffness variability.   We characterise the tendon unit and develop a finite element model that delineates its operation.  In a practical application, we integrate multiple tendon units into a bio-inspired jamming ankle mechanism, or {\em Jammkle}.  We experimentally show the benefits of the Jammkle in the context of robotic locomotion. 

This paper is structured as follows: next we frame the novel contributions of the work in the context of current related literature. In Section III we introduce the design of a tendon and describe the fabrication steps.  Section IV discusses characterisation and testing of the tendon.  Section V introduces the multiphysics model and demonstrates close agreement with experimental results.  In Section VI we describe the Jammkle and show the benefit of our design across a range of locomotion experiments.  In Section VII we discuss the results, and suggest extensions to the study.

\section{Background}
\subsection{Tendon Mechanisms}
Soft robotics harnesses flexible and adaptive materials to achieve unconventional, high-performance solutions and previously-unattainable capabilities through a combination of deformability, compliance, and variable stiffness \cite{shintake2018soft}.  Tendon mechanisms based on shape memory alloys \cite{lee2019long} and conventional strings have been applied to generate locomotion in arms \cite{mishra2017simba} and legs \cite{vikas2016design}.  These tendons are typically single 'string' elements rather than bundles, and are used for force transmission.  Here we focus on the compliance and damping of tendon mechanisms in a legged locomotion context.

\subsection{Soft Robotic Jamming \& Paws}
Jamming actuation is a popular vein of soft robotics research focused on producing tuneable-stiffness components through varying density of an actuation component under externally-applied stress, typically vacuum pressure.  Compared to other soft actuation mechanisms, jamming provides a rapid, dramatic stiffness variation with minimal volume variation \cite{fitzgerald2020review}.  Jamming also permits a wide variety of different actuator shapes, sizes and compositions \cite{howard2021shape,fitzgerald2021evolving}, allowing for flexible deployments in diverse applications \cite{Robertsoneaan6357}.

Jamming actuation comprises three main subfields: {\em granular} \cite{steltz2010jamming}, {\em laminar} \cite{1322995}, and {\em fibre} \cite{brancadoro2018preliminary} jamming, based on the type of actuation component used (membrane-enclosed groups of grains, stacked sheets of soft material, and bundles of flexible fibres respectively) \cite{fitzgerald2020review}.  

Granular jamming has been previously investigated in the context of rough-terrain locomotion, mainly through bag-style grippers mounted as robotic 'feet'.  Frictional and damping effects \cite{hauser2016friction} as well as high performance over deformable terrains has been noted \cite{chopra2020granular}.  Multi-material paws comprised of a jamming pad with individual toes, display high damping performance on gravel and rocky ground as well as acting as a gripper for manipulating objects \cite{howard2021one}.  The above designs are functionally predicated on direct contact between jamming bag and environment, which due to thin membranes required to maximise the jamming effect leads to wear and eventual failure (especially in rough terrain).    

\subsection{Fibre Jamming}
Fibre jamming is a nascent and relatively under-explored field.  Previous research has explored fibre jamming for single material fibres, assessing the impact of different fibre materials \cite{brancadoro2018preliminary} and arrangements \cite{brancadoro2020fiber}, and showing prototype systems for surgical manipulation \cite{brancadoro2019toward}. A continuum robot was created, demonstrating a large attainable stiffness range \cite{zhao2019soft}.

The closest work to our own uses 3D printed tendon bundles as a jamming end effector to grasp a range of objects \cite{jadhav2021variable}. The authors focus on single-material tendons and deflection properties of the manipulator. Additionally, the jamming structure is in direct contact with the environment.  Finally, the work does not consider the use of properties such as compliance and damping, nor use in legged locomotion.  

\subsection{Summary}
To summarise, our work progresses the state of the art in fibre jamming.  The main novel contributions of our work are:

\begin{itemize}
    \item Development of a versatile fibre jamming tendon unit that can be deployed in various locations on a robot.
    \item The use of multi-material printing \cite{howard2021one,joyee2019multi} to create heterogeneous fibre bundles for fibre jamming.
    \item The first research that considers fibre jamming as a solution for legged locomotion, removing the jamming mechanism from direct environmental contact for increased durability.
    \item Development of a finite element model that explains observed experimental behaviour.
\end{itemize}

\section{Tendon Design and Fabrication}

\begin{figure*}[h!]
\centering
\includegraphics[width=1.3\columnwidth]{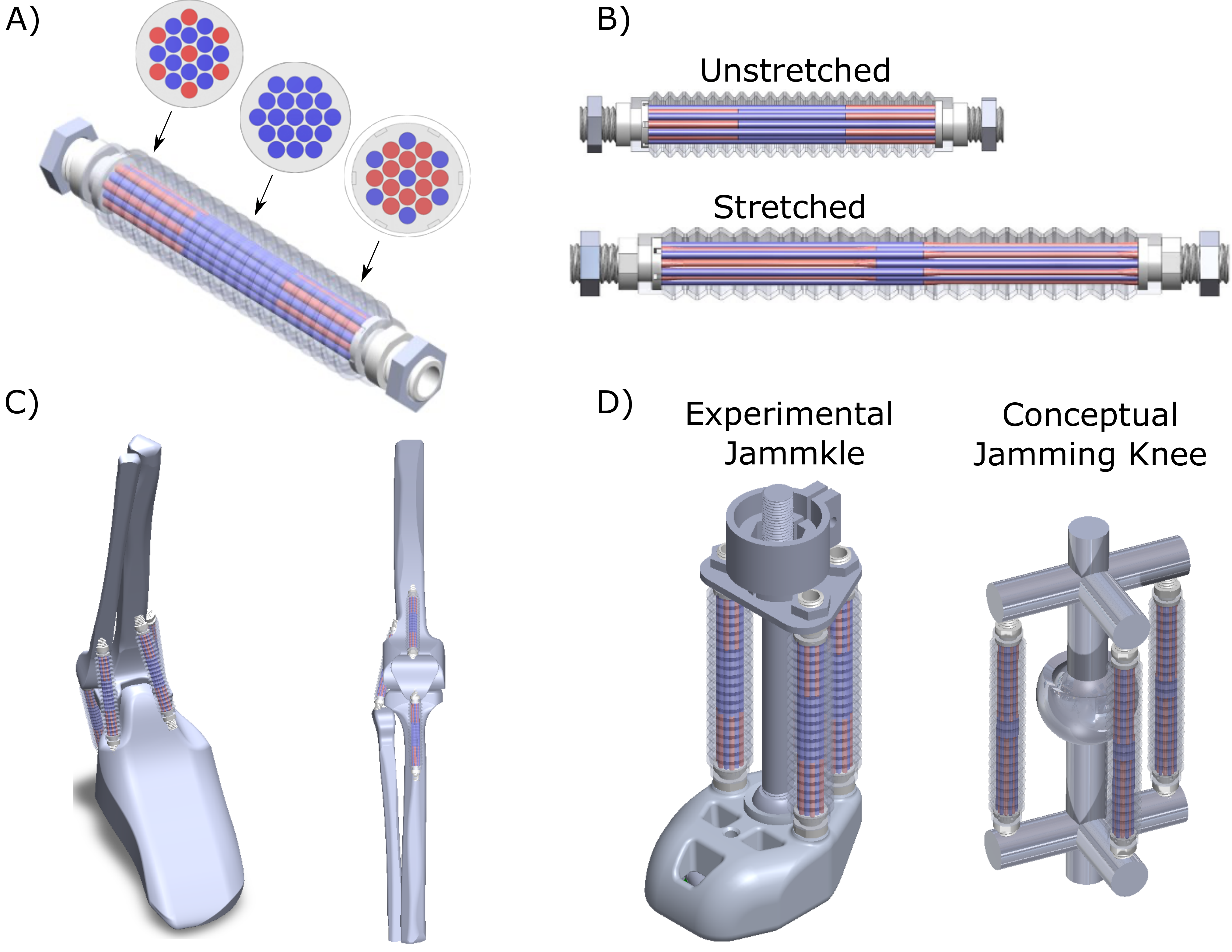}
\caption{Modular Design of Jamming Tendons: A) Jamming tendon showing cross-sections of three regions along its length. Short Shore A-30 fibres (red) connect to longer Shore A-85 fibres (blue). B) When stretched, extension occurs mainly in the softer Agilus30 fibres. C) The tendon is designed to be generically applicable and stabilise similarly to organic tendons; they can be modularly applied to joints including knees and ankles. D) Mechanical implementation of bio-inspired jamming ankle (Jammkle) and conceptual knee mechanism.}
\label{fig:fig2}
\end{figure*}

We develop a versatile tendon unit based on fibre jamming (Fig. \ref{fig:fig2}). As few design references for fibre jamming are available, we describe in detail; (1) the rationale and design behind the fibres and membrane, (2) the manufacturing method and material choices for the tendon, and (3) the testing method and characterisation of single tendons.. 

\subsection{Design of Fibres and Membrane}

A tendon is comprised of 18 dual-material cylindrical fibres, each \SI{2}{\milli\meter} in diameter, encased in an airtight elastomer membrane. The overall tendon is \SI{118}{\milli\meter} long, including two printed rigid (Shore D) end plates which contain mounts for vacuum tubing and are threaded for easy attachment to test equipment.  Each fibre is \SI{80}{\milli\meter} long, attaches to both end plates, and is comprised of two sections: a shorter (\SI{30}{\milli\meter}), highly deformable Shore A-30 section, and a slightly longer (\SI{55}{\milli\meter}), less deformable Shore A-85 section.  These sections and the tendon layout are shown in Fig. \ref{fig:fig2}.

 The fibres are arranged in a hexagonal tiling pattern for high surface contact and correspondingly high jamming force. The fibres are arranged such that every second fibre is mirrored about the centre of the tendon, with the longer, less elastic sections overlapped in the centre. The overlapping Shore A-85 sections halt the extension of the shorter sections and causes effective jamming under vacuum pressure. At atmospheric pressure the Shore-A 30 sections provide compliance.  The tendon's maximum extension before break is $\approx$\SI{55}{\milli\meter}.  

The external concertina membrane holds vacuum pressure and transfers the resulting force to the fibre bundle, pushing the fibres together to jam the tendon. The membrane is designed to have minimal effect on the function of the tendon.  We empirically justify this claim in Section \ref{SU_tests}.

\subsection{Tendon Fabrication}
The central section of the tendon units, including the fibres and solid threaded ends, were 3D printed in a single piece using a Stratasys Connex3 Objet500 polyjet printer.  3D printing enables the design as multimaterial segments of the tendon which are intended to be joined can be printed as a single piece, whilst tendon segments which are intended to move independently are printed with SUP706 soluble support material separating them.  The membrane surrounding the tendon is printed separately, as unrestricted physical access is required to fully remove support material from between the fibres. Once support material is removed, the tendon is assembled by stretching the membrane over the fibre bundle.

Polyjet printing allows us to tune Shore hardness through material mixing.  Our tendon uses three materials: firstly a rigid material (Vero, Shore D) which is used for the end plates.  The fibre sections are printed from soft Agilus30 (Shore A-30), and a mixture of Vero and Agilus30 which provides Shore A-85. Agilus30 is chosen for relatively high durability, high flexibility, high elongation at break, and high tensile tear resistance.  The Shore-A 85 materials balances favourable properties, e.g., sufficient compliance to bending, whilst retaining minimal elongation. Material properties for the two base resins are summarised in Table \ref{tab:material_data}.

\begin{table}[]
    \centering
        \caption{ Material Data for 3D Printed materials}
    \begin{tabular}{c|ccc}
&    \multicolumn{2}{c}{\bf Material }  \\
{\bf Property}   &Vero & Agilus30\\
\hline
Tensile Strength &20 - \SI{65}{\mega\pascal} &2.4 - \SI{3.1}{\mega\pascal}\\
Elongation at Break & 10-\SI{20}{\percent} & 220 -  \SI{270}{\percent} \\
Shore Hardness &83 - 86 Scale D &30 - 35 Scale A\\
Tensile Tear Resistance & - & 5 -  \SI{7}{\kilogram\per\centi\meter} \\

    \end{tabular}

    \label{tab:material_data}
\end{table}

\section{Tendon Characterisation and Testing}
Tensile testing assessed the mechanical properties of the tendon unit.  Tests were performed using a custom testing rig consisting of a linear actuator that controls movement and speed in a fixed vertical axis using a lead screw (see Fig. \ref{fig:fig3}). The test rig is equipped with a load cell (Zemic H3-C3-25kg-3B) and a 9 axis Microstrain 3DM-Dx5-25 IMU\footnote{Only vertical measurements are used.} .  The tendon is secured to the test rig via printed threads. One end is secured to a fixed plane, and the other is secured to the load cell which is moved at a controlled speed to capture the deformation.  The fixed end of the tendon is connected to a Thomas 107CDC20 vacuum pump via silicone tubing.

\subsection{Test Procedure}

\begin{figure}[h!]
\centering
\includegraphics[width=0.9\columnwidth]{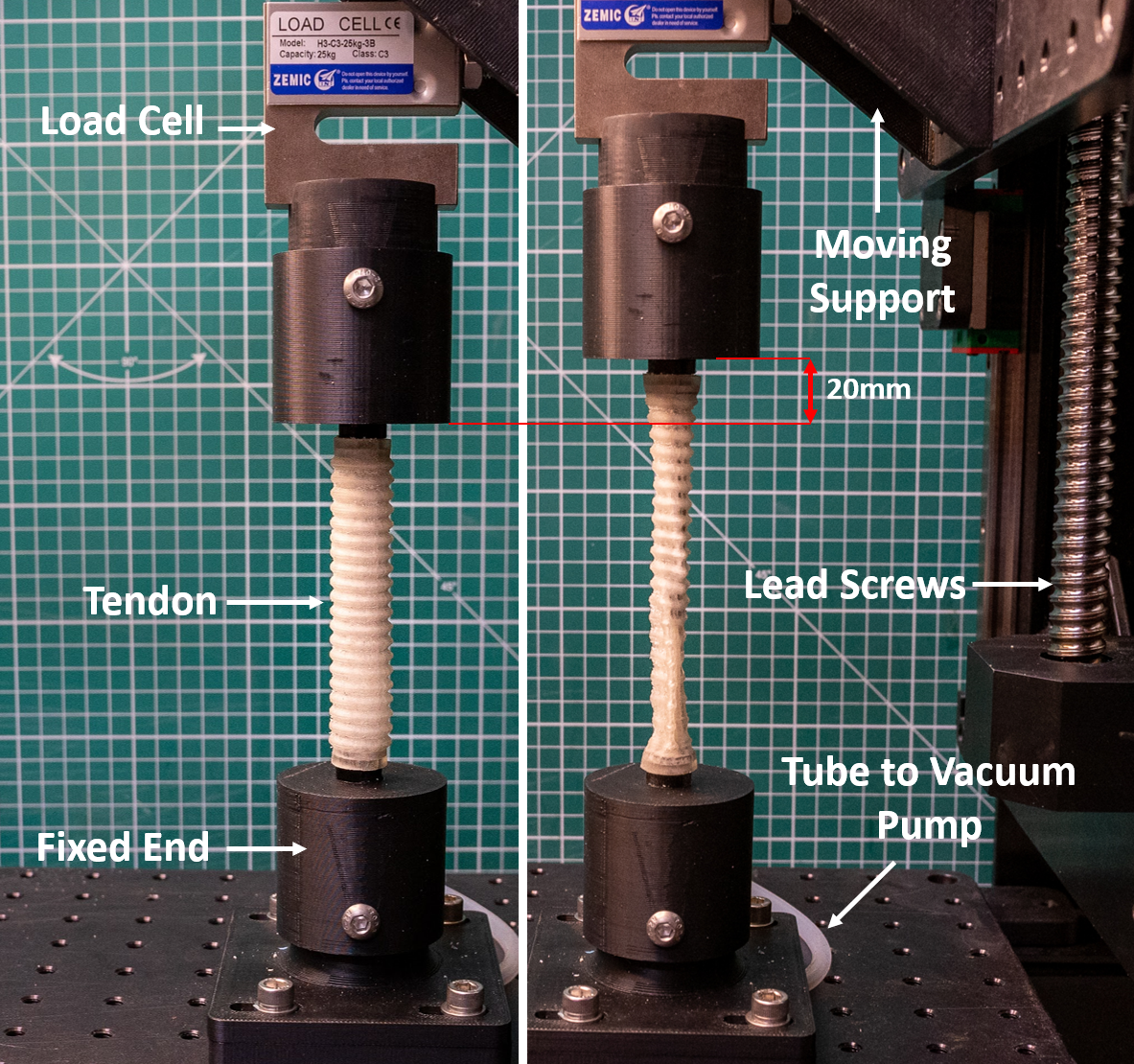}
\caption{Test setup used for characterisation of a single Tendon unit. Left panel shows unjammed, unstretched tendon. Right panel shows jammed, stretched tendon.}
\label{fig:fig3}
\end{figure}

The tendon is initially under no extension or vacuum. From this position, the movable support rises vertically at a rate of \SI{5}{\milli\meter\per\second} until \SI{20}{\milli\meter} extension is reached. The force required for the change in extension is measured by the load cell and force curves are recorded. The movable support then returns to the initial position.  If tendons are jammed as part of the test, the vacuum pump applies \SI{-50}{\kilo\pascal} pressure before the support is moved (right pane, Fig. \ref{fig:fig3}), and is released after the support returns.  All tests are repeated 5 times.  

\subsection{Tendon Characterisation}
\label{SU_tests}
The jamming tendons were characterised by sequentially altering the tendon's material composition and jamming pressure, each in isolation. Results are detailed in Table \ref{tab:characterisation_tests}.

The fibre bundle was tensile tested without the external membrane to measure the effect of the membrane on tendon functionality (Table \ref{tab:characterisation_tests} rows 1-2).  Results show mean peak force values for tendons tested with and without a membrane are within standard error.  In other words, the membrane does not have a significant effect on tendon performance. This justifies our modelling approach (Section \ref{sec:modelling}).
 
To compare the efficacy of the multi-material fibres to previously-studied single material fibre jamming, tests were run on tendons using single-material Shore A-30 fibres, and single-material Shore A-85 fibres (Table \ref{tab:characterisation_tests} rows 3-4).  Single material fibres represent the current state of the art in fibre jamming, e.g., \cite{jadhav2021variable}.  The multi material fibre tendons exhibited a \SI{112.5}{\percent} increase in mean peak force between unjammed and jammed (\SI{-50}{\kilo\pascal}) states.  Comparatively, the Shore A-30 fibre tendons exhibited an increase of \SI{27.8}{\percent}, and the Shore A-85 fibre tendons increased \SI{15}{\percent} when jammed. Results indicate that the combination of two material types within a single fibre shows a clear increase in the tendon's tunable compliance range for the same pressure range.

\begin{figure}[h!]
 \centering
 \includegraphics[width=0.9\columnwidth]{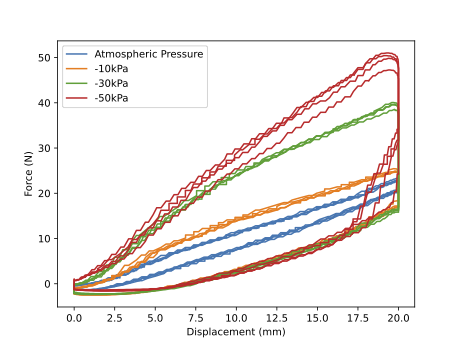}
 \caption{Experimental Force vs Displacement curves of multimaterial tendon. The effective stiffness of the tendon is able to be controlled by vacuum pressure.  Five repeats of each test are shown.}       
 \label{fig:JammingPressurePlots}
 \end{figure}

The vacuum pump used in this study can reliably maintain \SI{-50}{\kilo\pascal} vacuum pressure on the tendon.  To characterise the effect of vacuum pressure on tendon performance, the tests are repeated on the multimaterial tendon with \SI{-10}{\kilo\pascal} and \SI{-30}{\kilo\pascal} vacuum pressure (Table \ref{tab:characterisation_tests} rows 5-7).  The mean peak force results show an initial slight increase in force from atmospheric to \SI{-10}{\kilo\pascal}, and then a relatively linear increase in max peak force when vacuum is ramped from \num{-10} to \SI{-50}{\kilo\pascal}, demonstrating the tendon's controllable stiffness (Fig. \ref{fig:JammingPressurePlots}). It is expected that the effective stiffness could be increased with a more powerful vacuum pump, however larger pumps are unsuitable for mounting on a mobile legged robot.

\begin{table}[]
    \centering
    \caption{Results from Characterisation Tests performed on Tendons.* denotes our developed tendon in standard operating conditions (\SI{-50}{\kilo\pascal} vacuum, membrane, multi-material). Atm = atmospheric pressure, Mm = multi-material fibres.}
    \begin{tabular}{p{3.2cm}|p{1.4cm}p{1.4cm}p{0.9cm}}
&    &  {\bf Test Result }    &\\
{\bf Test type}   &Max Peak Force (\si{\newton}) & Mean Peak Force (\si{\newton}) & Std. err\\
\hline
No membrane, Atm. &26.125 & 23.634 & 0.663 \\
With membrane, Atm. &24.879 & 23.330 & 0.406 \\
\\

Shore A-30, Atm. &9.110 & 8.997 & 0.038 \\
Shore A-30, Vac \SI{-50}{\kilo\pascal} &11.601 & 11.501 & 0.032 \\
Shore A-85, Atm &92.133 & 84.294 & 1.993 \\
Shore A-85, Vac \SI{-50}{\kilo\pascal} &110.119 & 96.492 & 3.535 \\
\\

Mm, Vac \SI{-10}{\kilo\pascal} &27.292 & 25.415 & 0.488 \\
Mm, Vac \SI{-30}{\kilo\pascal} &41.423 & 39.852 & 0.466 \\
Mm, Vac \SI{-50}{\kilo\pascal}* &52.534 & 50.214 & 0.864 \\

    \end{tabular}
    \label{tab:characterisation_tests}
\end{table}

\section{Jamming Unit Modelling}
\label{sec:modelling}
The behaviour of the jamming tendon is modelled via FEM in COMSOL Multiphysics (e.g., \cite{collins2021review}).  Relative stiffness of the tendon in the jammed and unjammed state is evaluated using a 1/6th model, exploiting the tendon's radial symmetry to reduce the model to a \SI{60}{\degree} slice about its central axis.

The simulation scenario replicates the procedure used for tensile testing, with tendons stretched to \SI{20}{\milli\meter} in both jammed and unjammed states. We consider only a forward model (increasing displacement), with viscoelasticity negated to enable a tractable model. The membrane is modelled as a linear spring. To jam the tendons, \SI{-50}{\kilo\pascal} pressure is applied directly to the outer tendons.

Material properties of the Shore D Vero material were obtained directly from the manufacturer.  Stiffness of the Shore A-30 and Shore A-85 materials are experimentally determined through a 'dogbone' ASTM D638-14 uniaxial tensile test \cite{ASTM2016}, and stress-strain curves identified experimentally. An incompressible hyperelastic Yeoh model is fitted to the Shore A-30 data, resulting in Yeoh parameters $C_1=\SI{1.2e-2}{\mega\pascal}$,  $C_2=\SI{1.0e-4}{\mega\pascal}$, $C_3=\SI{6.2e-3}{\mega\pascal}$. Owing to its negligible nonlinearities, the Shore A-85 material is modelled as linearly elastic with modulus $E=\SI{7.34}{\mega\pascal}$.

Coefficients of friction for the three material pairings (A-30/A-30, A-30/A-85. and A-85/A-85) are estimated by fitting the parameters to experimental jamming data. A non-linear least squares fit is performed using the trust-region-reflective algorithm \cite{Branch1999}. At each iteration, the tendon model is jammed to $\SI{-50}{\kilo\pascal}$ then stretched to $\SI{20}{\milli\meter}$. A cubic polynomial is then fitted to the simulated force displacement curve to enable interpolation/extrapolation of data points to increase the robustness of the solver to convergence idiosyncrasies.

\begin{figure}[h!]
\centering
\subfloat[]{
\includegraphics[width=0.9\columnwidth]{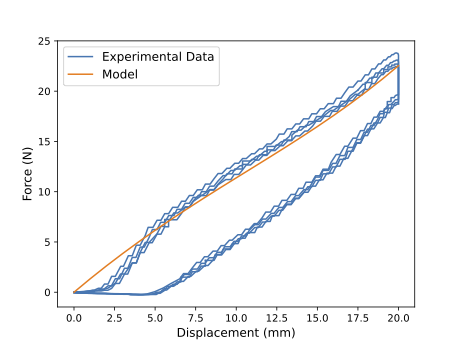}}\\
\subfloat[]{
\includegraphics[width=0.9\columnwidth]{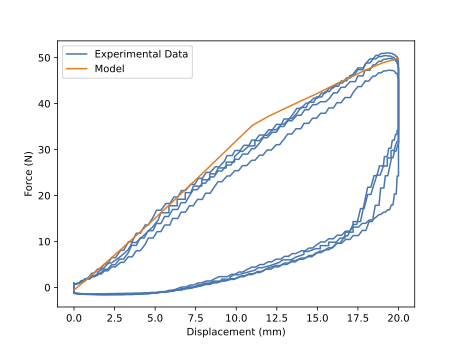}}
\caption{Force-displacement curves showing modelled and experimental behaviour of the jamming tendon during elongation: (a) unjammed, (b) jammed.}
\label{fig:SimulatedJamming}
\end{figure}

The coefficients of friction for the three material parings are estimated as $\mu_{A30-A30}=0.74$, $\mu_{A30-A85}=0.75$, and $\mu_{A85-A85}=0.99$, respectively. The simulated force-displacement curves for the jammed and unjammed tendons are presented in Fig. \ref{fig:SimulatedJamming}, and the model is shown to accurately capture the tendon's behaviour. 

\begin{figure}[h!]
 \centering
 \includegraphics[width=0.9\columnwidth]{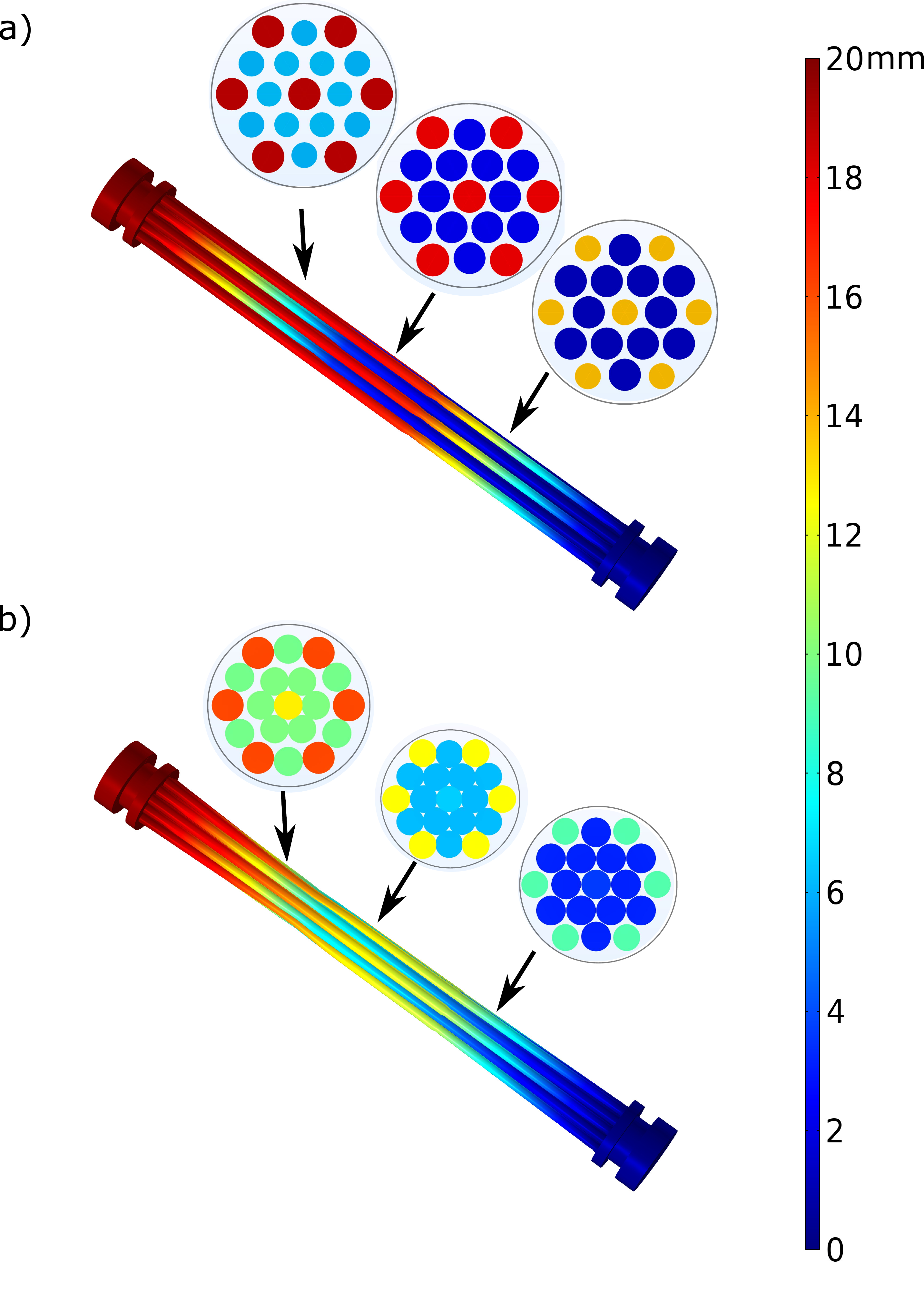}
 \caption{Modelled deformation of tendons during tensile testing, including cross sectional deformations. A \SI{20}{\milli\meter} displacement is applied to both the (a) unjammed and (b) jammed tendons. Without jamming, deformation is localised in the Shore A-30 fibre sections, however under jamming it is distributed between the two materials.}       
 \label{fig:ComsolFig}
 \end{figure}

The model provides insights into the function of the multi-material tendon.   Simulated deformation of the \SI{20}{\milli\meter} stretched jammed and unjammed tendon is presented in Fig. \ref{fig:ComsolFig}, showing cross-sectional deformations along its length. Unjammed, deformation arises from elongation of the compliant Shore A-30 fibre sections.  
When jammed, inter-fibre friction prevents the stiffer Shore A-85 segments from sliding across each other and forces them to stretch instead. Hence, the total deformation is distributed approximately uniformly along the length of the tendon, reducing the significance of the orientation of the A-30 and A-85 materials within each fibre. 

\section{Jammkle Assembly and Testing}

\begin{figure}[h!]
\centering
\includegraphics[width=0.77\columnwidth]{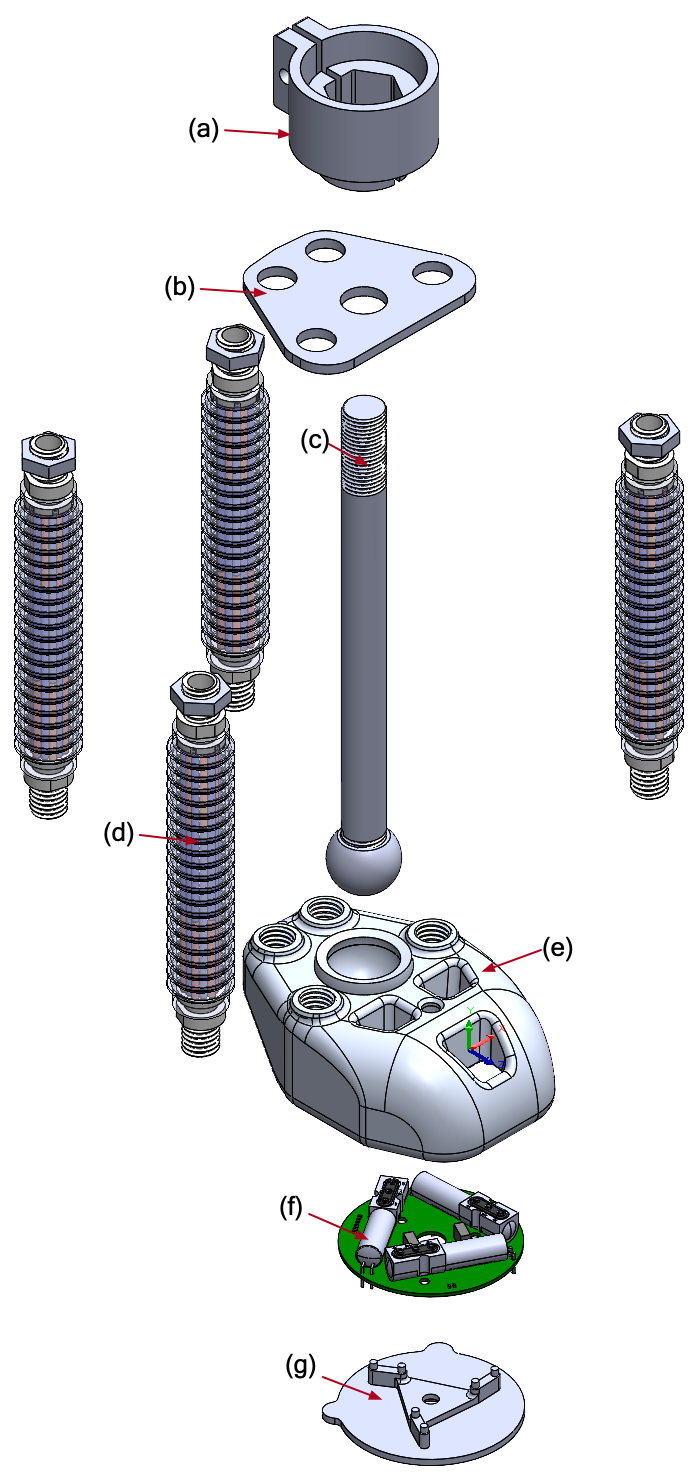}
\caption{An exploded model of the Jammkle assembly, including (a) mounting bracket, (b) tendon alignment plate, (c) rigid ball screw, (d) tendon, (e) printed foot, (f) pneumatic distribution board, (g) board mounting plate.}
\label{fig:fig7}
\end{figure} 

In this section we demonstrate the effectiveness of our tendon in a practical application - a bio-inspired ankle for legged locomotion over rough-terrain (Fig. \ref{fig:fig7}). We test the compliance and slip characteristics of this ankle, and benchmark it against another commonly utilised robotic feet.

\subsection{Construction}
The Jammkle is inspired by an abstracted human ankle. A central ball socket printed directly into a "foot" acts as an analog of a bone ankle joint. This socket allows for \SI{360}{\degree} axial rotation of the foot, which is limited by the tendons arranged around the joint. Tendons on either side of the foot are placed similarly to peroneal tendons and stabilise the ankle, preventing inwards and outwards rolling. Two tendons at the rear are placed to represent an achilles tendon. As the achilles tendon typically experiences more force than the interior stabilising tendons, two tendons were placed instead of one, to increase load bearing capabilities. A silicone tube is routed away from the ankle to the vacuum pump.

The foot was printed in rigid Vero material, which effectively transferred force to the tendon structure instead of deforming in contact with the terrain. 

A modular pneumatic distribution board comprised of 3 pneumatic solenoids controlled the state of each tendon (jammed or unjammed).  The board interfaces with pneumatic distribution ports printed directly into the foot and routed to the tendons, negating the need for external tubular routing.  The entire jammkle assembly weighs \SI{352}{\gram}.

\subsection{Experiments and Results}

\begin{figure}[h!]
\centering
\includegraphics[width=0.98\columnwidth]{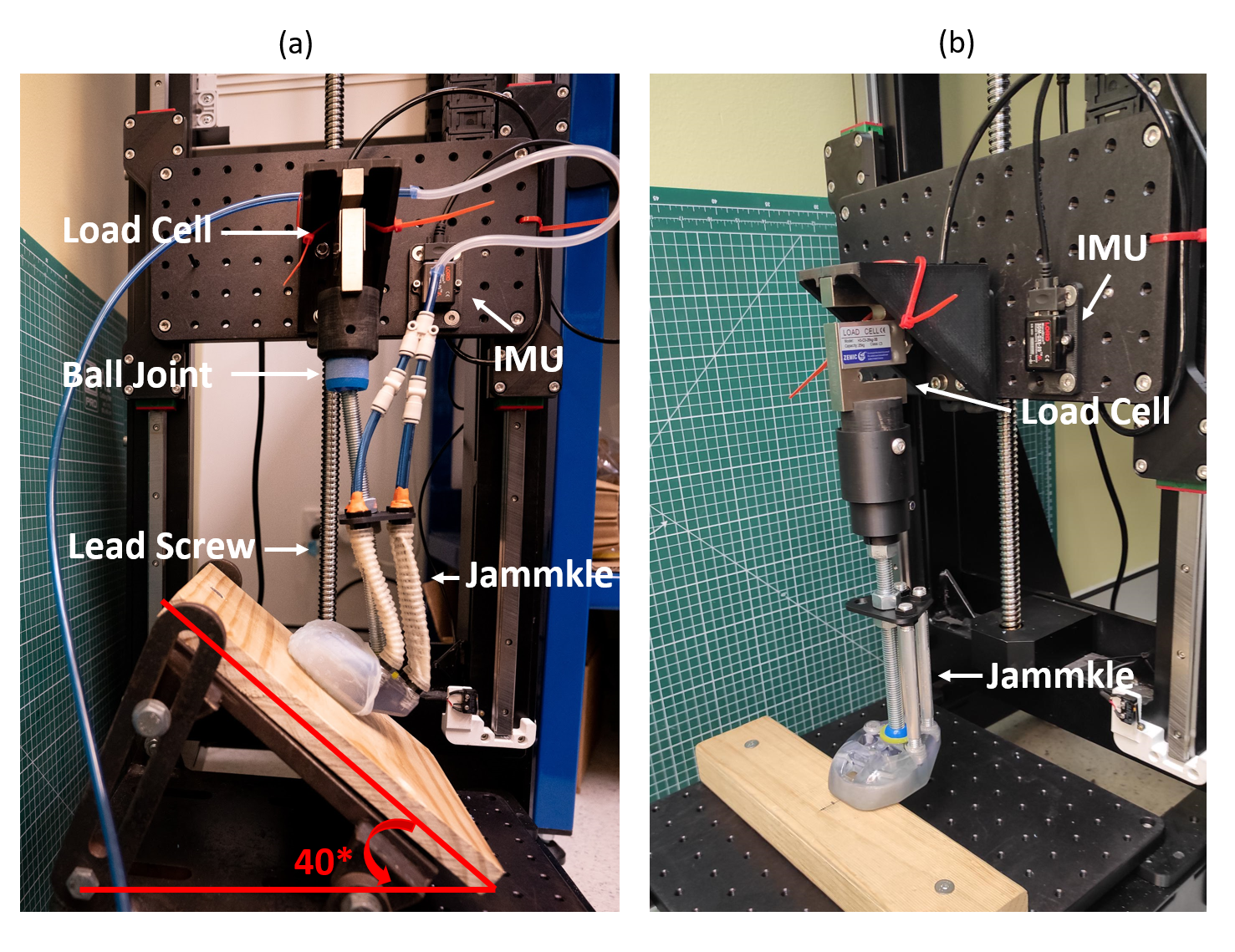}
\caption{The experimental setup for (a) slip test on a \SI{40}{\degree} slope (b) drop test.}
\label{fig:fig_setup}
\end{figure}

\subsubsection{Drop Test}
We demonstrate the ankle's damping and shock absorption properties under dynamic loading. A critically damped response with low maximum deceleration is desirable, decreasing the shock to legged components.

The test setup is demonstrated in Fig. \ref{fig:fig_setup}. The tests were performed using the same testing apparatus as previously. The linear actuator was decoupled from the platform, so that there was no resistance against the platform's freefall. The IMU was mounted to the test platform. The platform was then dropped from \SI{50}{\milli\meter} and the maximum negative acceleration in the vertical direction was measured by the IMU. The entire platform assembly weighed \SI{1.8}{\kilogram}. The test was repeated 5 times and the results averaged.
The test was repeated for 5 different benchmark feet: 
\begin{itemize}
    \item \textbf{Rigid Ankle:} An ankle constructed like the Jammkle but with solid aluminium tendons.  This allows us to isolate performance contributions from the tendons from contributions from the foot design.\
    \item \textbf{Standard Leg:}A rigid carbon-fibre leg with a 3D printed plastic tip.\
    \item \textbf{Squash Ball:} A rigid carbon-fibre leg with a squash ball attached to the bottom.\
    \item \textbf{Jammed Ankle:} The Jammkle with vacuum applied to the tendons.\
    \item \textbf{Unjammed Jamkle:} The Jammkle without vacuum applied to the tendons.\
\end{itemize}

\begin{figure*}[ht!]

\begin{center}
\centering 
\subfloat(a){\includegraphics[width=0.62\columnwidth]{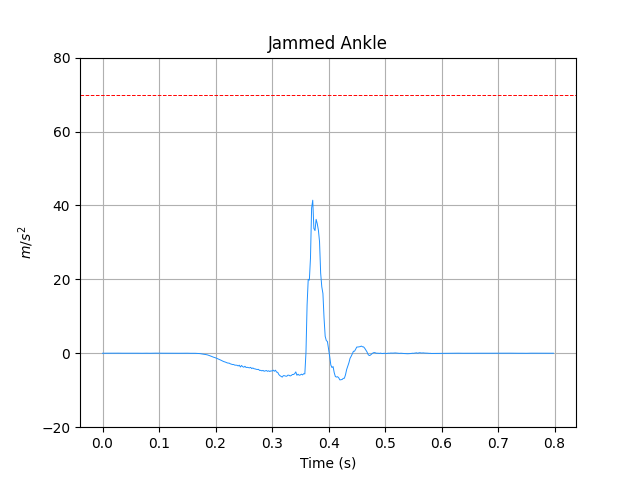} }
\subfloat(b){\includegraphics[width=0.62\columnwidth]{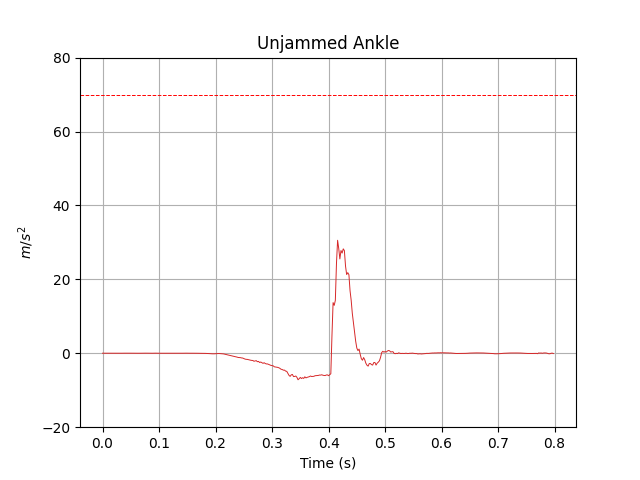}  }
\subfloat(c){\includegraphics[width=0.62\columnwidth]{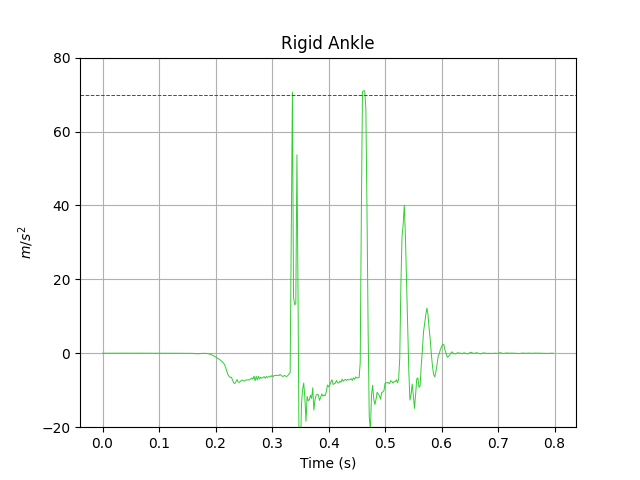}  }\\
\subfloat(d){\includegraphics[width=0.63\columnwidth]{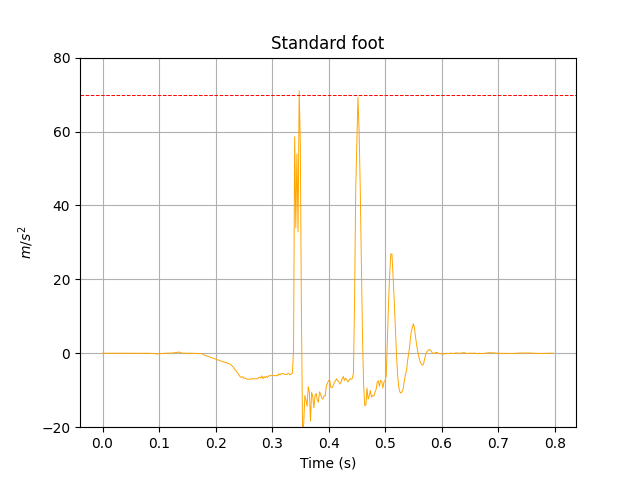}  }
\subfloat(e){\includegraphics[width=0.63\columnwidth]{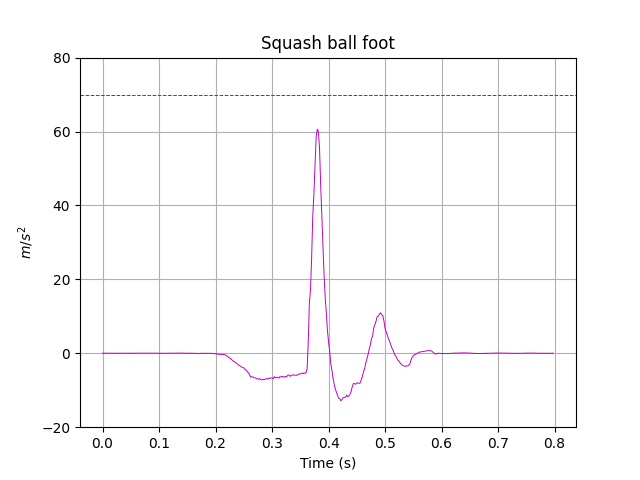}  }
\end{center}
\caption[]{Drop test results, showing critical damping of acceleration for the jammed and unjammed Jammkle.  Larger spikes are evidenced for the comparative feet, with the standard foot and rigid ankle bouncing significantly.}
\label{fig:fig9}
\end{figure*}

The results are summarised in Table \ref{tab:drop_test}.  Results show a clear difference in shock absorption. The unjammed Jammkle performed best, with a maximum deceleration of \SI{32.6}{\meter\per\second\squared}. The unjammed ankle also exhibited less bouncing and provided a highly damped response (see Fig. \ref{fig:fig9}).

\begin{table}[]
    \centering
        \caption{Drop test results showing negative acceleration (\si{\meter\per\second\squared}) for five comparative foot structures: Rigid ankle, standard robotic leg, squash ball, jammed Jammkle, and unjammed Jammkle. * denotes the sensor limit was reached.}
    \begin{tabular}{c|ccc}
&    &  {\bf Drop Test }    &\\
{\bf Comparative structure}   &Max & Mean & Std. err\\
\hline
Rigid Ankle &71.040* & 71.023 & 0.005 \\
Standard Foot &71.022* & 71.002 & 0.005 \\
Squash Ball Foot &61.359 & 60.602 & 0.230 \\
Jammed Ankle &43.187 & 42.520 & 0.314 \\
Unjammed Ankle &32.621 & 31.259 & 0.451 \\
    \end{tabular}

    \label{tab:drop_test}
\end{table}

\subsubsection{Slip Test}
Slipping is an undesirable occurrence in legged locomotion, as it requires the software control system to react quickly and can cause instability and falls. The slip test provides data on the forces the ankle can apply to sloped terrain before frictional force is overcome. The data gained provides insight into both static and kinetic frictional forces for the comparative ankles.  
The test utilises the same testing apparatus before with the addition of a wooden slope --- see Fig. \ref{fig:fig_setup}. The top of the ankle is attached to a ball joint, allowing the tibia angle to increase as the ankle slides down the slope. 
The ankle is then moved vertically downward at a rate of \SI{5}{\milli\meter\per\second} toward a piece of plywood inclined at \SI{40}{\degree}. After \SI{110}{\milli\meter}, the descent is halted. The test is repeated 5 times for each of the comparative ankles.

\begin{figure*}[ht!]
\begin{center}
\centering 
 \subfloat(a){\includegraphics[width=0.93\columnwidth]{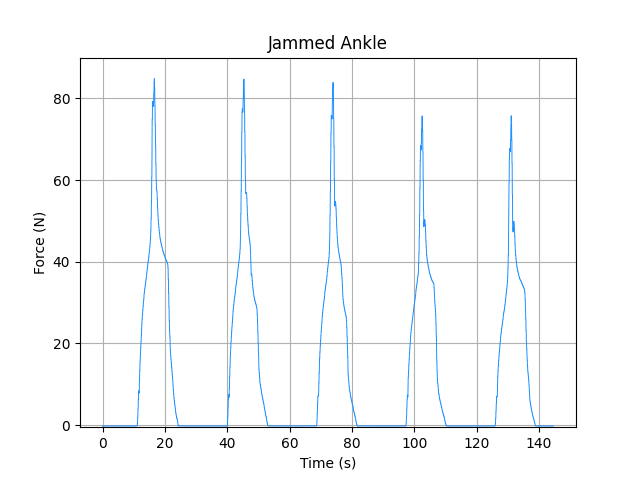} } 
 \subfloat(b){\includegraphics[width=0.93\columnwidth]{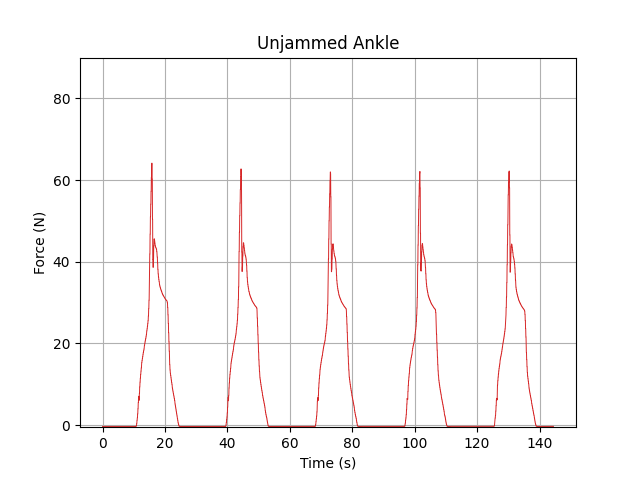}  }\\
 \subfloat(c){\includegraphics[width=0.93\columnwidth]{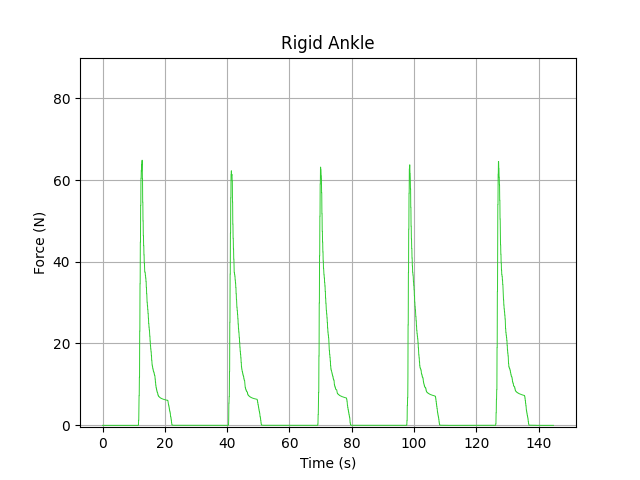}  }
 \subfloat(d){\includegraphics[width=0.93\columnwidth]{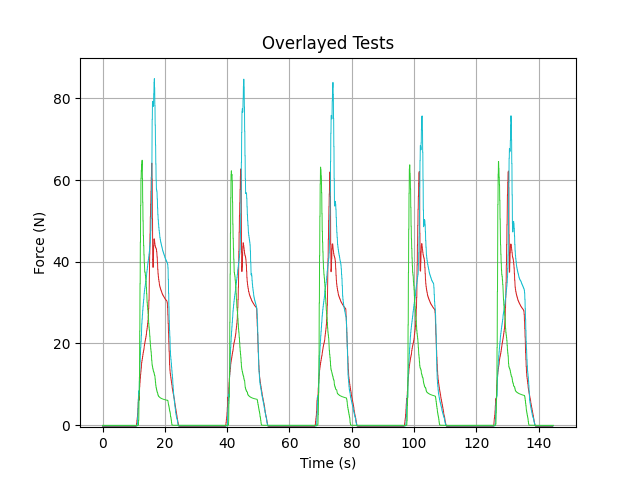}  }
\end{center}
\caption[]{Slip test results.  The foot is placed on a \SI{40}{\degree} slope and downwards pressure applied. Larger force dissipation equates to more slip resistance.}
\label{fig:fig8}
\end{figure*}

The jammed ankle had the largest peak force with both the unjammed and rigid ankle producing similar maximum force outputs before sliding (Fig. \ref{fig:fig8}). Fig. \ref{fig:fig8}(d) shows that the total force dissipated by the jammed  ankle is significantly greater than the rigid ankle structure, leading to reduced slip and correspondingly more stable footing for slope traversal when attached to legged platforms.

\section{Discussion}

To summarise, we demonstrated a soft robotic 3D printed multi-material fibre jamming tendon, and highlighted the advantages of multi-material tendons in achieving high stiffness variability.  We subsequently developed a finite element model that shows close agreement with experimental data and provides a mechanism for the functioning of the tendon.  Finally, we deployed the tendon into a jamming ankle mechanism, and showed a powerful practical application in providing damping and compliance for legged locomotion wherein the compliant element is removed from direct environmental contact to improve durability.  The Jammkle was shown to outperform comparative foot mechanisms, as well as being individually controllable and easily fabricated.

We have not yet fully exploited the independently-controllable nature of the tendons when distributed across robotic structures, nor explored their use as active actuators.  Future work will focus on these two research avenues, as well as integration of model-based optimisation as a part of larger morphology discovery frameworks, e.g., \cite{chand2021multi}.

Multimaterial printing provides new scope for the application of fibre jamming in a soft robotics field, providing distinctly different performance regimes compared to previous fibre jamming examples from the literature.  Tuning to specific cases can be easily achieved through tuning material properties within the fibres.  Because of this, we believe multi-material fibre jamming to be widely applicable across soft robotics where compliance, damping, and high stiffness variability are required.

\bibliographystyle{ieeetr}
\bibliography{references}

\end{document}